%% file: main.tex
\documentclass[10pt, conference, compsocconf]{IEEEtran}

\IEEEoverridecommandlockouts

\ifCLASSINFOpdf
\else
\fi
\let\labelindent\relax
\usepackage[numbers]{natbib}
\usepackage{graphicx}
\usepackage{times}
\usepackage{epsfig}
\usepackage{threeparttable}
\usepackage{amsmath}
\usepackage{textcomp}
\usepackage{hyperref}
\usepackage{import}
\usepackage{graphics}
\usepackage{array}
\usepackage{booktabs}
\usepackage{caption}
\usepackage{subcaption}
\usepackage{xcolor}
\usepackage{enumitem}
\usepackage{tabularx}
\usepackage{multirow}
\usepackage{stfloats}
\usepackage{amssymb}
\usepackage{latexsym}
\usepackage{pbox}
\usepackage{multicol}
\usepackage{tikz}
\usepackage{float}
\usepackage[export]{adjustbox}
\usepackage[framemethod=TikZ]{mdframed}
\usepackage{colortbl}

\begin{document}
%
\title{ICDAR 2019 Competition on Scene Text Visual Question Answering\\}

\author{\IEEEauthorblockN{
Ali Furkan Biten\textsuperscript{$\dagger, 1$}\thanks{\textsuperscript{$\dagger$}Equal contribution}, 
Rub\`{e}n Tito\textsuperscript{$\dagger, 1$}, 
Andres Mafla\textsuperscript{$\dagger, 1$},\\ 
Lluis Gomez$^1$, 
Mar\c{c}al Rusi{\~n}ol$^1$, 
Minesh Mathew$^2$, C.V. Jawahar$^2$, 
Ernest Valveny$^1$,  
Dimosthenis Karatzas$^1$}

\IEEEauthorblockA{$^1$Computer Vision Center, UAB, Spain \\
$^2$CVIT, IIIT Hyderabad, India \\
\tt\small \{abiten, rperez, amafla, lgomez, marcal, ernest, dimos\}@cvc.uab.es}
}

%


\maketitle
\begin{abstract}
This paper presents final results of ICDAR 2019 Scene Text Visual Question Answering competition (ST-VQA). 
ST-VQA introduces an important aspect that is not addressed by any Visual Question Answering system up to date, namely the
incorporation of scene text to answer questions asked about an image.
The competition introduces a new dataset comprising $23,038$ images annotated with $31,791$ question / answer pairs where the answer is always grounded on text instances present in the image.
The images are taken from $7$ different public computer vision datasets, covering a wide range of scenarios.

The competition was structured in three tasks of increasing difficulty, that require reading the text in a scene and understanding it in the context of the scene, to correctly answer a given question. A novel evaluation metric is presented, which elegantly assesses both key capabilities expected from an optimal model: text recognition and image understanding.

A detailed analysis of results from different participants is showcased, which provides insight into the current capabilities of VQA systems that can read. We firmly believe the dataset proposed in this challenge will be an important milestone to consider towards a path of more robust and general models that can exploit scene text to achieve holistic image understanding.
\end{abstract}


\section{Introduction}
Visual Question Answering (VQA) has grown into a popular research area in the Computer Vision community, as evidenced by a series of recent works ~\cite{antol2015vqa}, \cite{gao2015you}, \cite{ren2015exploring}, \cite{goyal2017making}, \cite{johnson2017clevr}, ~\cite{agrawal2018don}. Objective of VQA is to answer a natural language question asked about an image.

A considerable percentage of images, especially those taken in urban environments, contain text. Textual information appearing in the scene carries explicit, important semantic information that more often than not is necessary in order to fully understand the scene. Many real-life visual question answering cases (see for example the VizWiz challenge\footnote{\url{http://vizwiz.org/data/}}), are frequently grounded on scene text present in the scene. But despite the popularity of VQA systems, 
integrating the rich semantics of scene text in VQA systems
has not been explored to date.

Leveraging scene text information in a VQA scenario implies a shift from existing models that cast VQA as a classification problem, to generative approaches that are able generate novel answers (in this by recognizing and integrating scene text as necessary in the answer).




For the proposed "Scene Text Visual Question Answering" (ST-VQA) challenge, 
we employ a new dataset, introduced by organizers of the challenge~\cite{biten-tito-mafla-2019scenetextvqa}. The questions and answers in this dataset are defined in such a way that no question can be answered without reading/understanding scene text present in the given image.

Interestingly, concurrently with the ST-VQA challenge, a work similar to ours introduced a new dataset~\cite{singh2019} called Text-VQA. This work and the corresponding dataset were published while ST-VQA challenge was on-going. Hence we had no  opportunity to present a comparison of the two works/datasets in this edition of the the challenge report.


\section{Competition Protocol}
The ST-VQA Challenge ran between February and April 2019. Participants were provided with a training set at the beginning of March, while the test set images and questions were only made available for a two week period between 15-30 April. The participants were requested to submit results over the test set images and not executables of their systems. At all times we relied on the scientific integrity of the authors to follow the established rules of the challenge.

The Challenge was hosted at the Robust Reading Competition (RRC) portal\footnote{\url{https://rrc.cvc.uab.es/?ch=11}}. The RRC portal was developed in 2011 to host the original robust reading competitions concerning text detection and recognition from born-digital and scene images and has since evolved to a fully-fledged platform for hosting academic contests. At the time of running this challenge, the portal hosts $14$ different challenges, structured in $45$ different tasks. The platform currently has more than $10,000$ registered users from over $100$ countries, with more than $36,000$ methods evaluated to date.
The results presented in this report reflect the state of submissions at the closure of the official challenge period. The RRC portal should be considered as the archival version of results, where any new results, submitted after this report was compiled 
will also appear. All submitted results are evaluated automatically, and per-task ranking tables and visualization options to explore results are offered through the portal.  
\begin{figure*}[t]
\begin{center}
\begin{tabular}{p{0.23\textwidth} p{0.23\textwidth} p{0.23\textwidth} p{0.23\textwidth}}
    \includegraphics[width=\linewidth,height=0.75\linewidth]{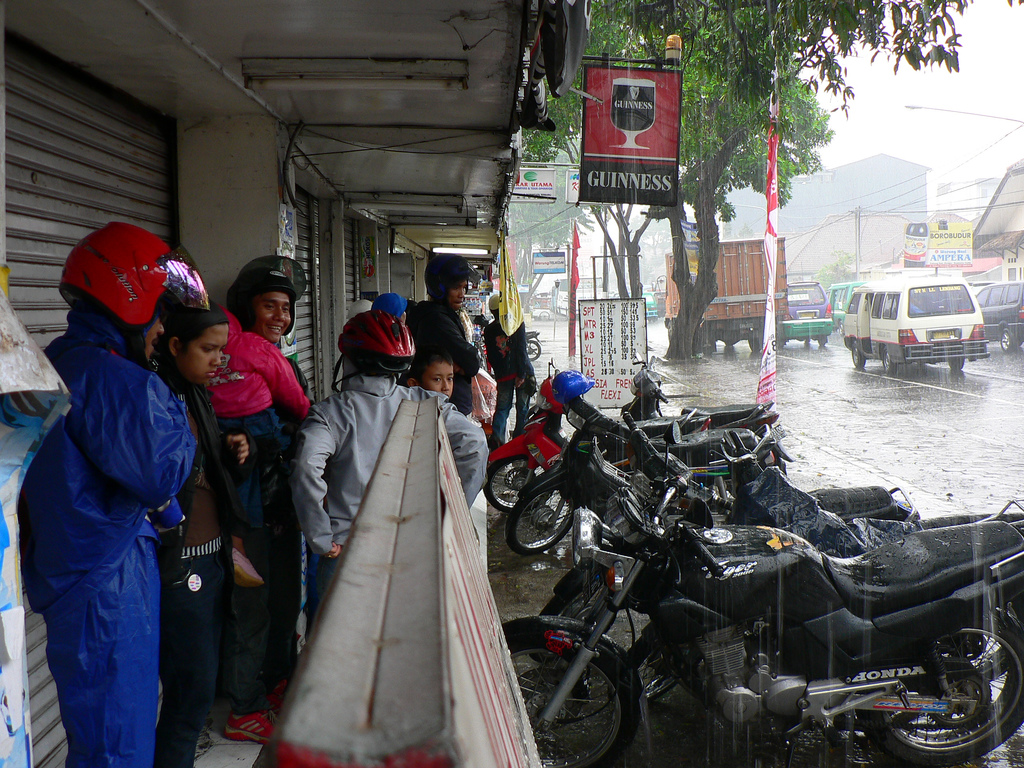} &
    \includegraphics[width=\linewidth,height=0.75\linewidth]{test_img_28.jpg} &
    \includegraphics[width=\linewidth,height=0.75\linewidth]{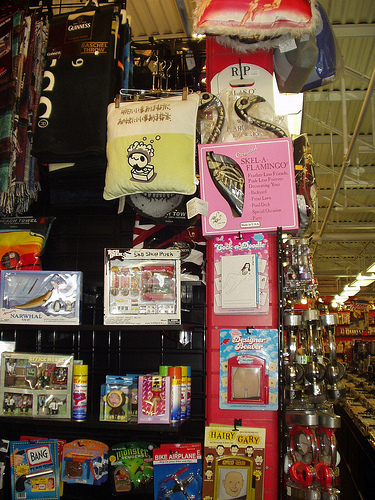} &
    \includegraphics[width=\linewidth,height=0.75\linewidth]{VizWiz_val_000000029853.jpg} \\
    \footnotesize{\fontfamily{qhv}\selectfont \textbf{Q:} What brand of alcohol is served at this establishment?} \par {\color{blue}\footnotesize{\fontfamily{qhv}\selectfont \textbf{A:} Guinness}} &
    \footnotesize{\fontfamily{qhv}\selectfont \textbf{Q:} What is the name of the library one of the signs is pointing to?} \par {\color{blue}\footnotesize{\fontfamily{qhv}\selectfont \textbf{A:} Lee Wee Nam Library}} &
    \footnotesize{\fontfamily{qhv}\selectfont \textbf{Q:} What is the name of the toy gun?} \par {\color{blue}\footnotesize{\fontfamily{qhv}\selectfont \textbf{A:} Bang}} &
    \footnotesize{\fontfamily{qhv}\selectfont \textbf{Q:} What is something found in the Bible?} \par {\color{blue}\footnotesize{\fontfamily{qhv}\selectfont \textbf{A:} The Ten Commandments}} \\
    & & &\\
    \includegraphics[width=\linewidth,height=0.75\linewidth]{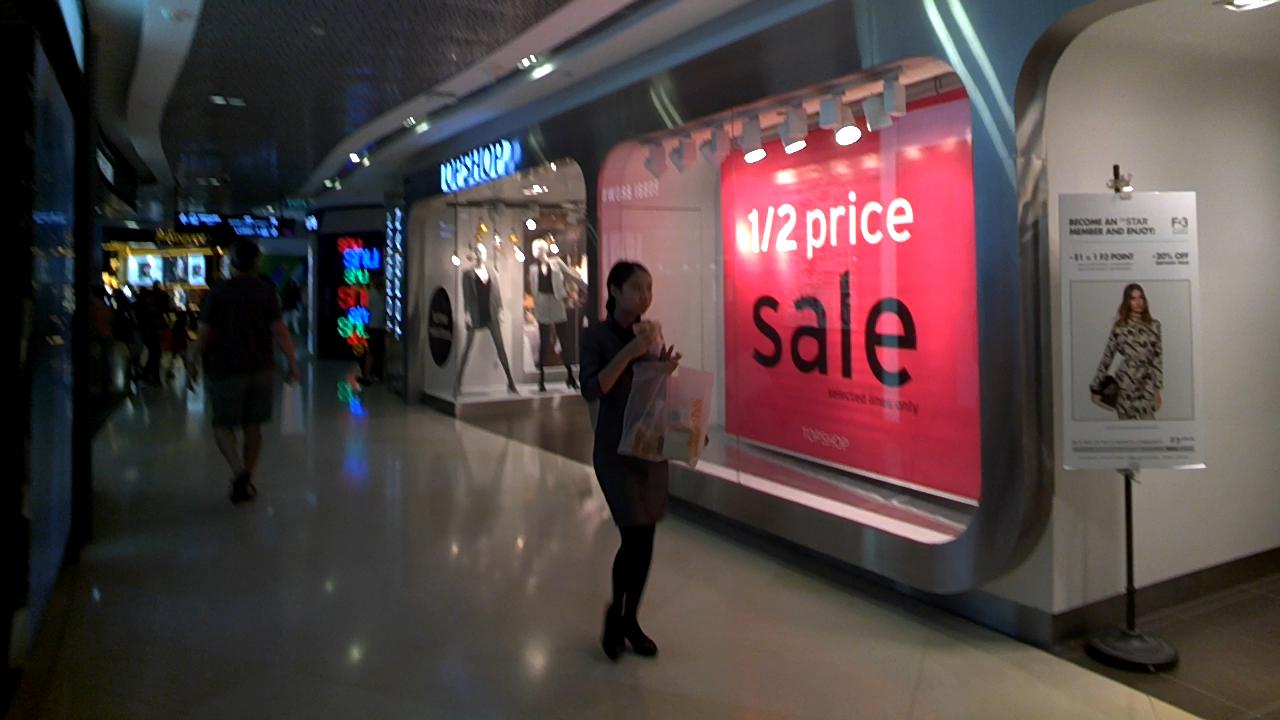} &
    \includegraphics[width=\linewidth,height=0.75\linewidth]{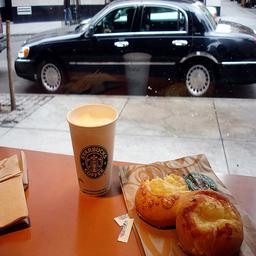} &
    \includegraphics[width=\linewidth,height=0.75\linewidth]{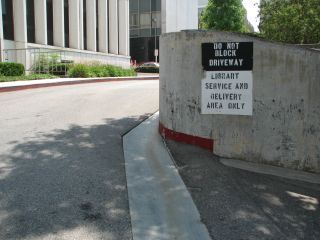} &
    \includegraphics[width=\linewidth,height=0.75\linewidth]{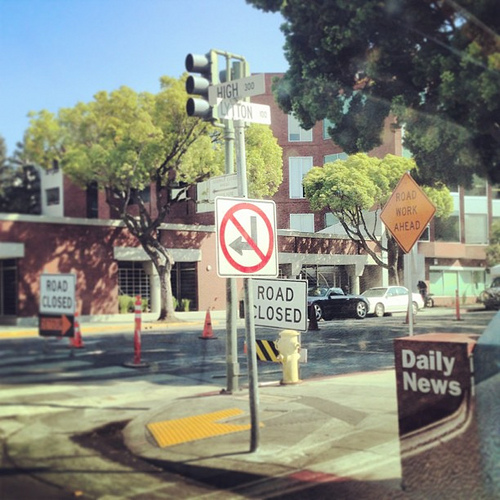} \\
    \footnotesize{\fontfamily{qhv}\selectfont \textbf{Q:} What word in black comes below 1/2 price?} \par {\color{blue}\footnotesize{\fontfamily{qhv}\selectfont \textbf{A:} sale}} &
    \footnotesize{\fontfamily{qhv}\selectfont \textbf{Q:} What company's logo is on the coffee cup?} \par {\color{blue}\footnotesize{\fontfamily{qhv}\selectfont \textbf{A:} STARBUCKS COFFEE}} &
    \footnotesize{\fontfamily{qhv}\selectfont \textbf{Q:} What is written in the black rectangle?} \par {\color{blue}\footnotesize{\fontfamily{qhv}\selectfont \textbf{A:} Do not block driveway}} &
    \footnotesize{\fontfamily{qhv}\selectfont \textbf{Q:} Which street sign is higher than the other?} \par {\color{blue}\footnotesize{\fontfamily{qhv}\selectfont \textbf{A:} HIGH}} \\
    & & &
\end{tabular}
\end{center}
\caption{Examples of questions and ground-truth answers from the ST-VQA training set}
\label{fig:main}
\end{figure*}

\section{The ST-VQA Dataset}
The ST-VQA dataset comprises 
images from seven different public datasets: ICDAR 2013~\cite{karatzas2013icdar}, ICDAR2015~\cite{karatzas2015icdar}, ImageNet~\cite{deng2009imagenet}, VizWiz~\cite{gurari2018vizwiz}, IIIT Scene Text Retrieval~\cite{MishraICCV13_IIIT_text}, Visual Genome~\cite{krishna2017visual} and COCO-Text~\cite{veit2016coco}. 
Sourcing images from different datasets 
reduces dataset bias (selection, capture and negative set bias) which popular computer vision datasets are subject to~\cite{khosla2012undoing}. In this case it also helps in obtaining better variability in questions and answers.
Since ImageNet, Visual Genome and VizWiz datasets do not have scene text annotations, we run a text retrieval model~\cite{gomez2018single} on these datasets to pick images containing text. Only those images on which the retrieval model found at least two text instances with high confidence, were retained.

The selected images were crowd sourced for annotation on Amazon Mechanical Turk (AMT).
We organized the annotation process in three steps. In the first step AMT workers were given images and specific instructions to come up with a question grounded on the text present in an image. It was mandated that the answer to the question must be text token(s) present in the image. 
In the second step a different set of workers were asked to give an answer given an image and the question(s) raised on the image in the first step. 
The collected answers were compared with the answers from the first step and any mismatch found resulted in passing the particular sample to a final verification step. 
At this stage, the ambiguous questions were checked by the organizers and corrected as necessary, before being added to the dataset. The final version of the dataset comprises $23,038$ images and $31,791$ question~/~answer pairs. 

\autoref{fig:main} shows some examples from our dataset. We appreciate the difficulty of the task, that requires not only to read the text correctly but also to understand the visual context in order to correctly answer the question. More details about the dataset can be found in \cite{biten-tito-mafla-2019scenetextvqa}.

\section{ST-VQA Challenge}
\label{sec:challenge}
The ST-VQA Challenge was structured as $3$ tasks of increasing difficulty. 

\textbf{Task 1 - Strongly Contextualized / Local Dictionaries:}
Following standard practice in word spotting tasks~\cite{karatzas2013icdar}~\cite{karatzas2015icdar}, we provide a dictionary of possible words related to the scene.
As such, we provide for each image a different dictionary of $100$ words that includes the correct answer among a number of distractors.
The distractors were generated using three methods.
Firstly we ran two reading systems \cite{gomez2018single,he2018end} on the image and output predictions above a certain threshold are added to the dictionary. 
Secondly additional words were added using  a method that generates contextualized lexicons for scene images using only visual information~\cite{patel2016dynamic}.
The remaining words were generated using regular expressions associated to the ground truths, thus producing similar words.
 
\textbf{Task 2 - Weakly Contextualized / Global Dictionary:}
In this task, a larger global dictionary, common for all images is provided. 
The global dictionary comprises $30,000$ words formed by collecting all the $22k$ unique ground truth answer words (from the training and test set) plus $8k$ words sampled from the individual dictionaries of task 1. This task is more challenging compared to Task 1, but on the other hand it still fits into the standard classification style VQA pipeline.

\textbf{Task 3 - Open Dictionary:}
The open dictionary task is the most generic and challenging one among all the the three tasks, since no dictionary is provided. To perform well on this task it would not be enough to follow the standard classification style modelling of the VQA problem where the answer is always one among a set of predefined classes. The models designed for this task should have ability to read the text present in the image and find the answer from the text tokens recognized from the image. 

Furthermore, we separate our test set for each task to a set of shared and specific images. Shared images are the ones that exist in all three three tasks, while specific images are unique to each task at hand. The number of shared images are $3,069$ in total while the number for specific images are around $500$ images. The logic behind such a division of the dataset is to compare the models within similar images while at the same time assess the models in a unique and diversified set of images across each task. 

\subsection{Evaluation Metric}
\label{seq:metric}
Standard accuracy based VQA evaluation metrics make a hard decision about correctness of the answer. This makes sense in case of classification pipelines, but is not suitable in case of the ST-VQA where  answer to a question is scene text recognized from the image. Hence we need an evaluation metric which responds softly to answer mismatches due to OCR imperfections.
Therefore we use average normalized Levenshtein distance~\cite{levenshtein1966binary}. Let \textrm{ANLS} refer to the average normalized Levenshtein similarity as defined in equation (\ref{eq:lev}), where $N$ is the total number of questions, $M$ the number of GT answers per question, $a_{ij}$ the ground truth answers where $i = \{0, ..., N\}$, and $j = \{0, ..., M\}$, and $o_{q_i}$ the returned answer for the $i^{th}$ question $q_i$. Then, the final score is as: 



\begin{equation}
\label{eq:lev}
\textrm{ANLS} = \frac{1}{N} \sum_{i=0}^{N} \left(\max_{j}  s(a_{ij}, o_{q_i}) \right) \\
\end{equation}

\[
  s(a_{ij}, o_{q_i}) =
  \begin{cases}
            (1 - NL(a_{ij}, o_{q_i})) & \text{if \textrm{ $NL(a_{ij},o_{q_i}) < \tau$}} \\
                                    $0$ & \text{if \textrm{ $NL(a_{ij},o_{q_i})$  $\geqslant$  $\tau$}} 
  \end{cases}
\]

\noindent
where $NL(a_{ij}, o_{q_i})$  is the Normalized Levenshtein distance between the strings $a_{ij}$ and $o_{q_i}$ (notice that the normalized Levenshtein distance is a value between $0$ and $1$). We then define a threshold $\tau = 0.5$ to filter NL values larger than this value by returning a  score of $0$ if the NL is larger than $\tau$. The intuition behind the threshold is that if an output has a normalized edit distance of more than $0.5$ to an answer, 
we reason that this is due to returning the wrong scene text instance, and not due to recognition errors. Otherwise, the metric has a smooth response that can gracefully capture errors both in providing good answers and in recognizing scene text. All methods submitted as part of the competition are evaluated automatically using the above protocol at the RRC portal. A stand-alone implementation of the evaluation metric can also be downloaded from the RRC portal\footnote{\url{http://rrc.cvc.uab.es/?ch=11}}.



\begin{table}
\begin{tabular}{m{1.15cm} m{7cm}}
\toprule
 Method & Description \\ \midrule
 VTA &  A similar model to Bottom-Up and Top-Down~\cite{anderson2018bottom} with BERT~\cite{devlin2018bert} encoding of question and text.\\\midrule
 USTB-TQA & Combination of object detection (OD), OCR, question representations to produce answers via performing an attention from OD representation to OCR representation.\\\midrule
 USTB-TVQA & Combination of image, question and OCR features to produce an answer.\\\midrule
  
 Focus & A similar model of Bottom-Up and Top-Down~\cite{anderson2018bottom} with open-ended answer generation\\\midrule
 VQA-DML &  Encoder-decoder architecture with n-gram output as an answer.\\\midrule
 TMT & A model similar to Dynamic Networks~\cite{xiong2016dynamic} with VGG-16~\cite{simonyan2014very}.\\\midrule
 QAQ & Simultaneous detection and recognition, sharing computation and visual information among the two complementary tasks.\\\midrule
 Clova-AI OCR & A model similar to MAC network~\cite{hudson2018compositional} with BERT~\cite{devlin2018bert} and pointing mechanism.\\
\bottomrule
\end{tabular}
\caption{Short descriptions of the participating methods to ST-VQA Challenge}
\label{tab:methods}
\end{table}

\section{Results and Analysis}


This section presents results of the submitted methods under each task along with their analysis. 
Final results for each task at the end of the competition period are provided in \autoref{tab:main}.
The results are reported using two evaluation metrics: ANLS -- the metric we have introduced in Section~\ref{seq:metric} -- and Accuracy, i.e. the percentage of questions for which the provided answer is exactly the same as the ground truth answer.
Presenting Accuracy measure along with the ANLS makes it possible for us to compare the ST-VQA challenge with the standard VQA setting where results are often reported in terms of classification accuracy.
We appreciate that current state-of-the-art VQA models achieve an accuracy around 70\% on the standard benchmark VQA v2.0~\cite{goyal2017making}, while the best accuracy score for task 2 of ST-VQA (closest in nature to standard VQA task) is 17\%, which illustrates the difficulty of the proposed challenge and dataset.


\subsection{Baselines}
We provide baseline results (for all tasks of the challenge) using two methods drawn from the recent scene text literature. Both baselines are designed to be question-agnostic, they ignore the questions and focus only on the scene text present in the image to provide an answer. Far from suggesting appropriate pipelines for this task, the rationale for providing such baselines is to explore the limits of ad-hoc question-agnostic approaches.

The first baseline is based on the scene text retrieval method presented in~\cite{gomez2018single}, which jointly predicts word bounding boxes and a compact text representation of words given in a PHOC~\cite{almazan2014word} encoding. This baseline employs two criteria to come up with an answer. The first (``STR retrieval'') uses the given dictionary as a set of queries, and the top-1 retrieved word is taken as the answer. The second one (``STR largest''), returns the answer following the notion that humans tend to formulate questions about the largest text in the image. We sort the text found in an image by size, and the word that is contained in the largest bounding box is the answer chosen by the system.

For the second baseline we use a state of the art end to end scene text spotting model ~\cite{he2018end}. The detected text is ranked according to the confidence score obtained.
For tasks 1 and 2, the text candidate in the provided dictionary which best matches the most confident output is chosen as the answer. 
For task 3 the most confident output is directly adopted as the answer since no dictionary is provided.
\subsection{Submitted Methods}

Overall, 8 methods from 7 different participants were submitted for the $3$ proposed tasks in the ST-VQA challenge. All the methods followed an encoder-decoder architecture, which now is the de facto choice for Image Captioning and VQA problems.
Specifically, the submitted  methods are mostly based on the Bottom-Up and Top-Down attention model~\cite{anderson2018bottom} architecture. Additionally, most of the methods employ BERT~\cite{devlin2018bert} which is an off-the-shelf embedding method for encoding the questions or the text tokens predicted by an OCR model. A short description of each method can be found in \autoref{tab:methods}. 


\subsection{Task 1 - Strongly Contextualized/ Local Dictionaries}
In this task, 6 different methods have participated. The winning method is VTA both according to ANLS and accuracy, see \autoref{tab:main}. Although the first three methods performed significantly better than the baselines, the remaining three were below the two question agnostic  baselines. 

The difference between scores for Task 1 and other tasks is evidently due to the provided dictionary as explained in \autoref{sec:challenge}, suggesting that the methods took advantage of the specific dictionaries provided per image.

\input{joint_table.tex}
\input{answer_length.tex}

\begin{figure*}[t]
\begin{center}
\begin{tabular}{p{0.95\linewidth}}
\\\\
\includegraphics[width=\linewidth]{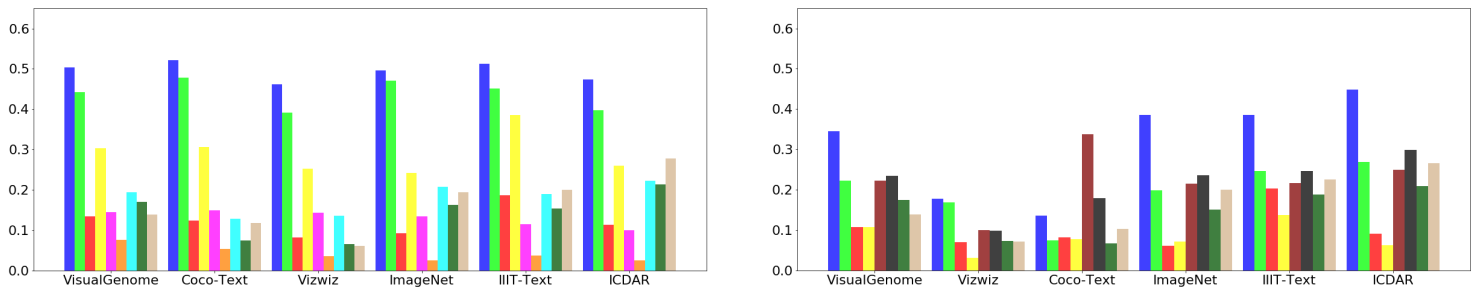} \\\\
\includegraphics[width=\linewidth]{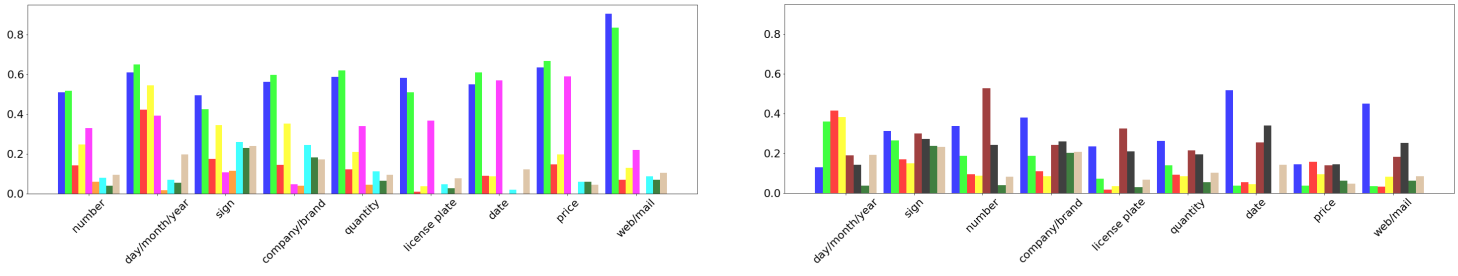} \\

\begin{center} \includegraphics[width=0.48\linewidth]{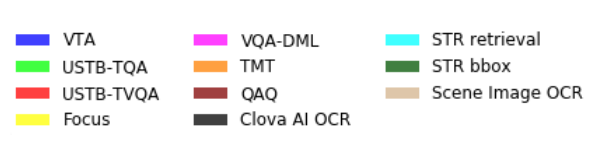} \end{center} \\
\end{tabular}
\caption{A detailed breakdown of the performance of the submitted models by image source (top) and question categories (bottom) for Task 1 (left) and Task 3 (right).}
\label{fig:len&source}
\end{center}
\end{figure*}

\subsection{Task 2 - Weakly Contextualized / Global Dictionary}
There are 4 submissions from 3 different participants in Task 2. Overall, we appreciate in \autoref{tab:main} a similar behavior of methods and baselines in this task and in Task 1, except for the expected general drop in performance since the dictionary provided in this case is not a local, smaller, scene specific dictionary.
VTA is again the best scoring method both according to ANLS and accuracy metrics. 

\subsection{Task 3 - Open Dictionary}
In the third task, there were $6$ submitted methods coming from $5$ different participants.
The best performing methods, like in case of the other two tasks is VTA. This shows the robustness of this method, although there is a considerable performance drop from task 1 to task 3.
In the results showcased in \autoref{tab:main}, the methods which participated in both Task 2 and Task 3, have a very similar performance.
Our conjecture for this phenomena is related to the size of the provided dictionary on Task 2. The significant size of the global dictionary acted as a distractor rather than as a guiding vocabulary to the models.


\subsection{Performance Analysis}
In this section we present an analysis of the performance of the submitted methods across the three tasks.

\textit{Shared vs Specific:}
The test set across the different tasks contains a shared amount of question/image pairs as well as a specific set defined for each task. This division of test sets allows us to assess the different models across all tasks in a generalized manner, while at the same time providing insight into the algorithms' performance on a unique, independent set of questions not available in other tasks.
It is worth noting that all the models perform in a similar manner in both the shared and specific sets. This result reinforces the veracity of two important assumptions: a) the division of the shared and specific sets capture a similar distribution of question and image types, and b) all the models use the provided dictionaries on task 1 and task 2. 

\textit{Question categories:}
In order to obtain better insight into each model's strengths, the performance according to different question categories is shown in \autoref{fig:len&source}. The question categories chosen cover the most common types of questions that refer to numbers, dates, sign, brands/companies, license plates, dates, prices, web/mails and quantities related to metric units. It becomes easy to spot that in task $1$, the questions regarding websites and e-mails are somewhat easier, since there are a lot of images that contain websites and mails that frequently refer to the photographers' information, thus creating a strong bias towards a specific answer. The hardest questions in task $1$ are the ones related to signs and license plates. Regarding signs question types, we believe this effect occurs because the answers require a specific understanding to select which sign the question refers to among all the detected OCR options. However, license plates questions are specific on the expected answer and contain a defined pattern. Nonetheless the low performance on models other than VTA or USTB-TQA most probably relate to issues on the OCR at reading license plates, specifically in COCO text images, in which license plates text is hard to spot and comes in very small scales.
On task $3$ the dates, prices and quantities question types show a significant decrease in performance. We infer that this is due to the large amount of available answer options for this type of questions when not using a strongly or weakly contextualized dictionary.

\textit{Analysis of Models' Output:}
Studying the models' output gives further intuition about the limitations of the methods. To this end, we show in \autoref{tab:answer_length} the score obtained for each method if we take into account only the model's answers with a specific length (number of words). Furthermore, we show the percentage of unique words/answer for each model and answer length to analyze the generative aspect of models as well as its ability to deal with out-of-vocabulary words. 

As can be observed from \autoref{tab:answer_length}, there is a clear drop in performance for all models in all tasks the answer length increases. Moreover, we observe that all the models except Clova and Focus strongly favors producing 1-word length answers, this might be because 60\% of the dataset answers are single words. Although Focus and Clova lag behind in performance, their distribution for answer length is quite similar to the ground truth. 

In order to know if the models use the provided strongly and weakly contextualized dictionaries, the percentage of answers that are contained in the dictionaries is shown in Table~\ref{tab:main}. The top performing method VTA employs the strongly contextualized dictionary on all the questions provided, the second method USTB-TQA uses it in $97.05$\% on the answers and the runner up method Focus employs it only in $68.84$\% of its answers, providing more than $1$K out of dictionary answers due to the generative nature of the model. However in Task $2$,  $48.91$\% of the answers submitted by VTA are contained in the weakly contextualized dictionary. The second and third methods (USTB-TQA and USTB-VTQA) have a similar percentage of answers that belong to the given dictionary, $84.11$\% and $83.76$\% respectively.
\begin{table}[h]
    \centering
    \begin{tabular}{l c c c c c }
        \toprule
         Methods&  1&  2& 3+&  Total& Vocab Size\\
         \midrule
         VTA& 45.22 & 10.14 & 4.01 & 59.40 & 2389 \\
         USTB-TQA& 32.08 & 0.00 & 0.00 & 32.10 & 1088 \\
         USTB-TVQA&34.64 & 0.03 & 0.00 & 34.70 & 1176\\
         Focus & 31.10 & 10.88 & \textbf{6.93} & 48.91 & 1998\\
         QAQ& \textbf{72.58} & 0.88 & 0.09 & 73.58 & \textbf{2495}\\
         Clova AI OCR & 43.57 & \textbf{20.99} & 6.01 & 70.61 & 2435 \\
         \midrule
         Ground Truth & 45.48 & 19.92 & 14.59 & 79.99 & 4596\\ 
         \bottomrule
    \end{tabular}
    \caption{Percentage (\%) of unique answers according to length and vocabulary size in Task 3.}
    \label{tab:unique}
\end{table}

To further investigate the generative power of models, we provide in \autoref{tab:unique} the percentage of unique answers for different answer lengths and their total vocabulary size (number of unique answers) for task 3. We see that even though VTA is the winning method, it is not as diverse as Clova or QAQ. Furthermore, we observe that most of the methods are not able to produce different answers for more than 3 tokens. However, QAQ and Clova get very close to ground truth in producing unique answers over the total set of answers. Finally, compared to ground truth, the vocabulary size of the models is quite limited. 

\begin{figure*}[ht!]
\begin{center}
\begin{tabular}{p{0.45\linewidth} p{0.45\linewidth}}
\includegraphics[width=\linewidth]{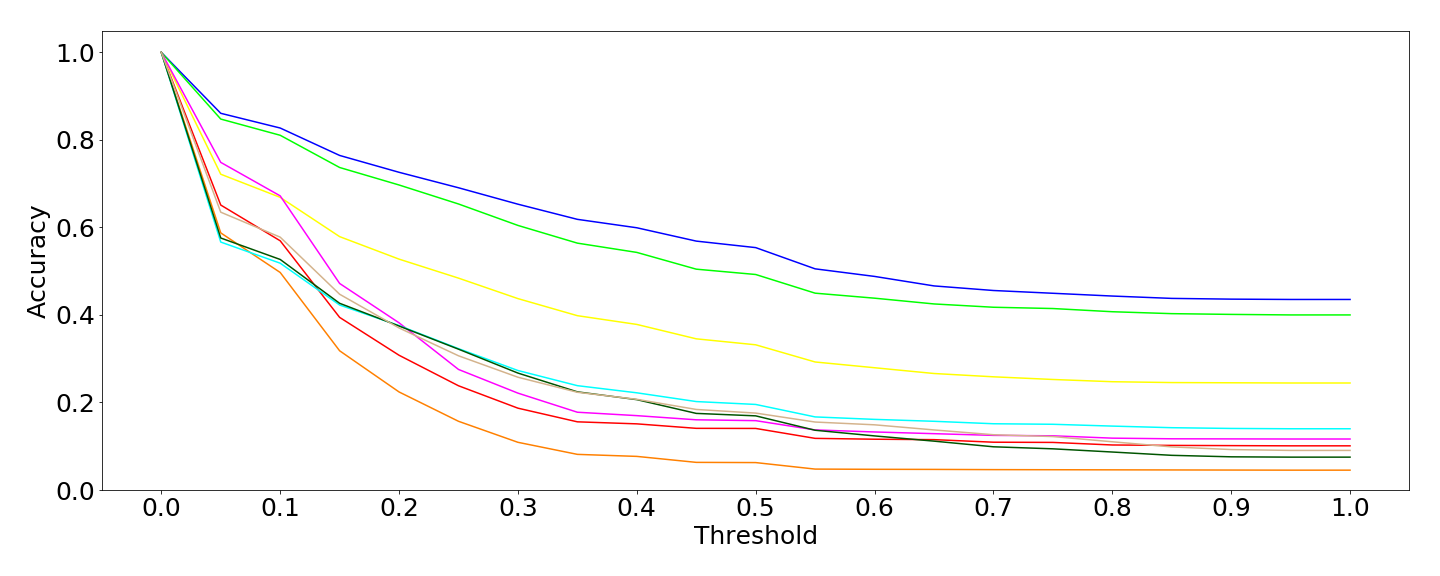} &
\includegraphics[width=\linewidth]{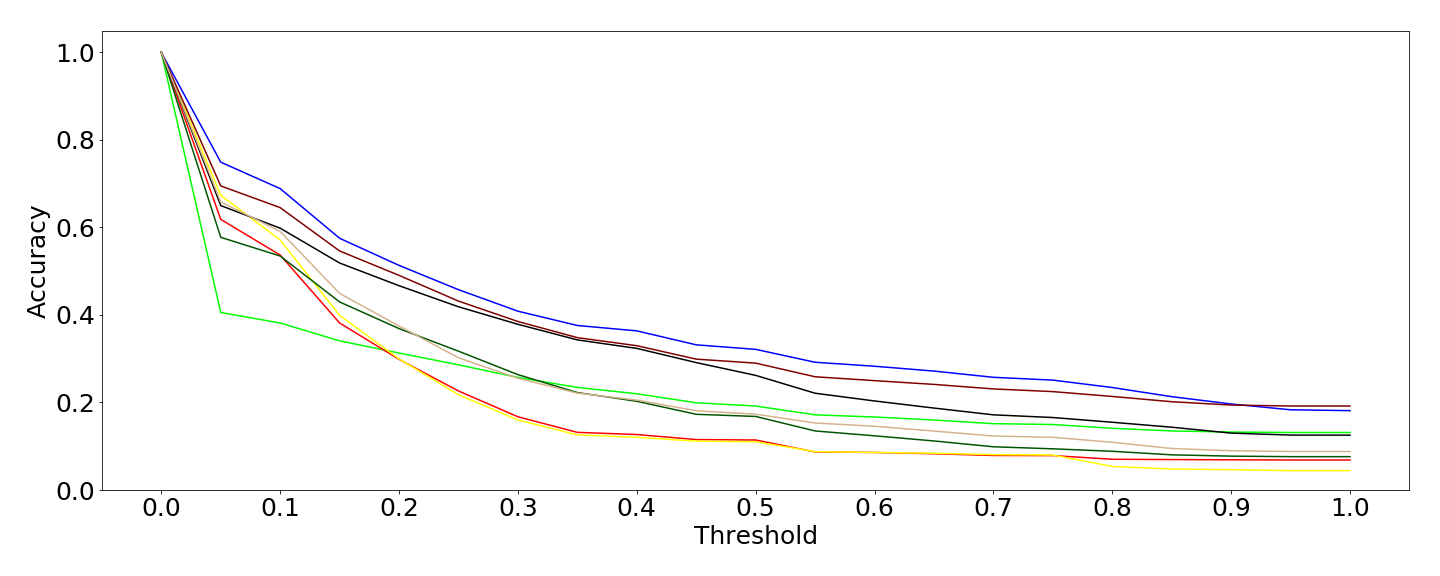} \\
\end{tabular}
\begin{center} \includegraphics[width=0.48\linewidth]{MethodsLegend.png} \end{center} 
\caption{Accuracy scores per ANLS threshold for Task 1 (left) and Task 3 (right)}
\label{fig:acc_threshold}
\end{center}
\end{figure*}
\textit{Results per dataset:}
It is a well-known fact in the literature that scoring high in one dataset does not necessarily translate to good performance in other datasets with the same task, since each dataset has its own biases and specific challenges that need to be addressed in a different way. 
To analyse this behaviour, we provide an analysis of the models' performance over images sourced from different datasets.

According to \autoref{fig:len&source}, VizWiz is the most challenging dataset while ICDAR and IIIT-Text are the easiest sources for both task 1 and task 3. This was expected since ICDAR and IIIT-Text tend to contain images where text is better focused and has larger size compared to the other datasets. At the same time images in VizWiz dataset are captured by visually impaired volunteers and hence the images are typically blurry, occluded and up-side town. 


One encouraging point is that for both task 1 and task 3, most of the models perform similarly on those datasets which are not curated for scene text detection and recognition problems: Visual Genome and ImageNet, suggesting that they can somewhat generalize to generic datasets not specifically collected for scene-text.

\textit{Interpretation of the ANLS metric:} Here we perform a analysis of the ANLS metric as a function of the threshold used to filter out wrong answers. We calculate the accuracy score according to the clipped ANLS score at different threshold values. To do so, we calculated the accuracy by accepting an answer as correct whenever its ANLS score is above the given threshold, see \autoref{fig:acc_threshold}. Contrary to the soft metric we used before, in this case we add 1 instead of adding 1-ANLS every time an answer is deemed to be correct. It can be noticed that the accuracy is quite stable for threshold values $> 0.5$ threshold, implying that the selected threshold in section \ref{seq:metric} is a good indicator of the model's performance. Additionally, we see once again that Task 3 is obviously more difficult than Task 1 by detecting the higher slope (sharper decrease) in the provided plots.

It is interesting to note that in this interpretation of the ANLS metric, for threshold value $\tau = 1.0$ the metric reverts to the classic Accuracy one. These plots provide therefore a good summary of the behaviour of the methods spanning the soft metric proposed here and the hard Accuracy typically used in VQA tasks.



\section{Conclusions and Future Work}
This work presents a novel VQA challenge in which the questions and answers are defined in such a way that no question can be answered without reading/understanding the scene text present in the images. The challenge is based on a new dataset with images collected from a wide range of sources, and question/answer pairs that have been collected according to the text found in them. In order to combine elegantly the recognition accuracy as well as the image understanding capacity of the participating methods, a new metric was proposed, namely Average Normalized Levenshtein Similarity. A thorough analysis of the different contestants' models has been provided. A breakdown of the results across the different source image dataset, answer lengths, and question categories is presented. The analysis provides insights of the strengths and weaknesses of each method. The results illustrate that the ST-VQA challenge is demanding, and will require future research methods to aim both towards increasing scene text recognition accuracy as well as moving towards full generative models in order to successfully tackle the proposed problem and similar tasks.

\section*{Acknowledgments}
This work has been supported by projects TIN2017-89779-P,
Marie-Curie (712949 TECNIOspring PLUS), aBSINTHE (Fundacion BBVA 2017), the CERCA Programme / Generalitat de Catalunya, a European Social Fund grant (CCI: 2014ES05SFOP007), NVIDIA Corporation and PhD scholarships from AGAUR (2019-FIB01233) and the UAB.

\bibliographystyle{IEEEtranS}
\bibliography{references}

\end{document}

%% file: joint_table.tex
\begingroup
\setlength{\tabcolsep}{5.23pt}
\begin{table*}[]
\scriptsize
\begin{tabular}{lccccccccccccccccc}
\toprule
                & \multicolumn{5}{c}{Task 1}                             && \multicolumn{5}{c}{Task 2}                             && \multicolumn{4}{c}{Task 3}                 &  \\ 
                \cmidrule{2-6} \cmidrule{8-12} \cmidrule{14-17}
Method          & Shared & Specific & Total          & Acc.  & Dict. && Shared & Specific & Total          & Acc.  & Dict. && Shared & Specific & Total          & Acc.  &  \\ \midrule
VTA             & 0.507  & 0.501    & \textbf{0.506} & \textbf{43.52} & 100       && 0.280  & 0.268    & \textbf{0.279} & \textbf{17.77} & 48.91     && 0.280  & 0.285    & \textbf{0.282} & 18.13 &  \\
USTB-TQA        & 0.457  & 0.445    & 0.455          & 39.98 & 97.05     && 0.168  & 0.196    & 0.173          & 13.34 & 84.11     && 0.168  & 0.183    & 0.170          & 13.14 &  \\
USTB-TVQA       & 0.129  & 0.100    & 0.124          & 10.09 & 20.55     && 0.093  & 0.094    & 0.093          & 6.59  & 83.76     && 0.093  & 0.108    & 0.095          & 6.86  &  \\
Focus           & 0.300  & 0.275    & 0.295          & 24.45 & 68.84     && 0.080  & 0.081    & 0.080          & 4.16  & 58.84     && 0.088  & 0.089    & 0.088          & 4.42  &  \\
VQA-DML         & 0.142  & 0.138    & 0.141          & 11.63 & 99.97     && -      & -        & -              & -     & -         && -      & -        & -              & -     &  \\
TMT             & 0.076  & 0.045    & 0.055          & 4.53  & 13.80     && -      & -        & -              & -     & -         && -      & -        & -              & -     &  \\
QAQ             & -      & -        & -              & -     & -         && -      & -        & -              & -     & -         && 0.255  & 0.265    & 0.256          & \textbf{19.19} &  \\
Clova AI OCR    & -      & -        & -              & -     & -         && -      & -        & -              & -     & -         && 0.213  & 0.224    & 0.215          & 12.53 &  \\ \midrule
STR \cite{gomez2018single} (retrieval) & 0.170  & 0.176    & 0.171          & 13.78 & 100       && 0.074  & 0.065    & 0.073          & 5.55  & 100     && -      & -        & -              & -     &  \\
STR \cite{gomez2018single} (largest)      & 0.126  & 0.134    & 0.130          & 7.32  & 100       && 0.120  & 0.107    & 0.118          & 6.89  & 100       && 0.125  & 0.142    & 0.128          & 7.21  &  \\
Scene Image OCR \cite{he2018end} & 0.146  & 0.145    & 0.145          & 8.89  & 100       && 0.134  & 0.125    & 0.132          & 8.69  & 100       && 0.137  & 0.154    & 0.140          & 6.60  &  \\ \bottomrule
\end{tabular}
\caption{Main Results Table. The columns `Shared' and `Specific' show the ANLS scores for the shared question/image pairs among the three tasks and for the specific set defined for each task respectively. ANLS score for all the samples taken together is shown in column titled `Total'. Accuracy metric for the entire set is shown in column, `Acc.'. The last column named `Dict.' lists the  percentage of output answers that are contained in the dictionaries provided for tasks $1$ and $2$.}
\label{tab:main}
\end{table*}
\endgroup

%% file: answer_length.tex
\begin{table*}[]
\centering
\begin{tabular}{lccccccccccccc}
\toprule
                & \multicolumn{6}{c}{Task 1}                                                                  &  & \multicolumn{6}{c}{Task 3}                                                                  \\ \cmidrule{2-7} \cmidrule{9-14}
                & \multicolumn{2}{c}{Length 1} & \multicolumn{2}{c}{Length 2} & \multicolumn{2}{c}{Length 3+} &  & \multicolumn{2}{c}{Length 1} & \multicolumn{2}{c}{Length 2} & \multicolumn{2}{c}{Length 3+} \\
Method          & Score          & Ratio  & Score          & Ratio  & Score           & Ratio  &  & Score          & Ratio  & Score          & Ratio  & Score           & Ratio  \\ \midrule
VTA             & \textbf{0.62}  & 79.08       & \textbf{0.31}  & 17.89       & 0.13            & 3.02        &  & 0.30           & 76.83       & \textbf{0.29}  & 18.08       & \textbf{0.19}   & 5.09        \\
USTB-TQA        & 0.60           & 95.56       & 0.16           & 4.22        & 0.02            & 0.22        &  & 0.22           & 100.00      & 0.07           & 0.00        & 0.01            & 0.00        \\
USTB-TVQA       & 0.17           & 99.88       & 0.01           & 0.12        & 0.00            & 0.00        &  & 0.13           & 99.95       & 0.01           & 0.05        & 0.00            & 0.00        \\
Focus           & 0.34           & 73.38       & 0.22           & 17.09       & \textbf{0.15}   & 9.53        &  & 0.11           & 72.33       & 0.06           & 17.40       & 0.03            & 10.27       \\
VQA-DML         & 0.19           & 98.03       & 0.03           & 1.00        & 0.00            & 0.98        &  & -              & -           & -              & -           & -               & -           \\
TMT             & 0.08           & 100.0       & 0.00           & 0.00        & 0.00            & 0.00        &  & -              & -           & -              & -           & -               & -           \\
QAQ             & -              & -           & -              & -           & -               & -           &  & \textbf{0.33}  & 98.82       & 0.11           & 1.03        & 0.02            & 0.15        \\
Clova AI OCR    & -              & -           & -              & -           & -               & -           &  & 0.24           & 66.61       & 0.19           & 25.92       & 0.10            & 7.47        \\ \midrule
Ground Truth    & 1.0            & 70.70       & 1.0            & 18.36       & 1.0             & 10.95       &  & 1.0            & 71.23       & 1.0            & 17.86       & -               & 10.91       \\ \bottomrule
\end{tabular}
\caption{Performance scores and ratio of the produced answers' length for Task 1 and Task 3. }
\label{tab:answer_length}
\end{table*}